\def\year{2019}\relax
\documentclass[letterpaper]{article} 
\usepackage{aaai19}  
\usepackage{times}  
\usepackage{helvet}  
\usepackage{courier}  
\usepackage{url}  
\usepackage{graphicx}  

\usepackage{amssymb}
\usepackage{multirow}
\usepackage{makecell}
\usepackage{float}
\usepackage[colorlinks,linkcolor=red]{hyperref}

\usepackage{}
\begin{document}
%
\title{Exploring Correlations for Multiple Facial Attributes Recognition through Graph Attention Network}
\author{Yan Zhang, Li Sun\\
School of Information Science and Technology\\
East China Normal University\\
Shanghai, China\\
}
\maketitle
\begin{abstract}
Estimating multiple attributes from a single facial image gives comprehensive descriptions on the high level semantics of the face. It is naturally regarded as a multi-task supervised learning problem with a single deep CNN, in which lower layers are shared, and higher ones are task-dependent with the multi-branch structure. Within the traditional deep multi-task learning (DMTL) framework, this paper intends to fully exploit the correlations among different attributes by constructing a graph. The node in graph represents the feature vector from a particular branch for a given attribute, and the edge can be defined by either the prior knowledge or the similarity between two nodes in the embedding with a fully data-driven manner. We analyze that the attention mechanism actually takes effect in the latter case, and utilize the Graph Attention Layer (GAL) for exploring on the most relevant attribute feature and refining the task-dependant feature by considering other attributes. Experiments show that by mining the correlations among attributes, our method can improve the recognition accuracy on CelebA and LFWA dataset. And it also achieves competitive performance.
\end{abstract}

\section{Introduction}
\noindent Facial image provides rich high level attributes which are useful for describing the semantics. Recognizing facial attributes has many real world applications in video surveillance \cite{vaquero2009attribute}, human-computer interaction \cite{cowie2001emotion} or image retrieval \cite{parikh2011relative}. Although, different attributes lie in distinct areas of facial regions, and may have different characteristics, recent works still tend to construct unified network to recognize them simultaneously \cite{liu2015deep,rudd2016moon,han2017heterogeneous,hand2017attributes}. The reasons are mainly as follows. First, it is costly in both time and space to build a deep network for each individual attribute. Second, recent results of deep learning shows that even totally different tasks actually share the same low level representation, therefore, both the structure and weights in lower layers can be shared among different tasks \cite{yosinski2014transferable}. Third, in multi-task learning (MTL), parameters of the network are optimized by minimizing the combined loss functions for each task. Thus, it is inherently easy to generalize \cite{meyerson2018pseudo}.

In general, the definition of MTL can be rather broad. As soon as there are more than one loss functions for optimizing, it is actually doing MTL \cite{ruder2017overview}. Tasks in MTL may even not have the same data during training, but still only one single model is available for making multiple predictions in testing phase. Specifically, this paper considers multiple facial attributes recognition. In this topic, each facial image in the training set is labeled with multiple binary attributes, such as male, young, brown hair, eye glasses \emph{etc.}, and our goal is to design a network through which we can obtain multiple predictions for attributes. Similar to previous works, we also share lower layers with the purpose of mining correlations among different attributes, and make branches to learn the feature representation for each unique attribute at higher level until it gives the final binary results. In such a framework, the different task correlations are only reflected in lower layers. Once branches separate, they are considered to be independent of each other in later layers. Thus, there are no sharing of feature in higher layers, which means that task related correlation is not exploited enough. We argue that sharing lower layers, which even happens between two irrelevant tasks in transfer learning, is obviously inadequate for these relevant tasks. 

\begin{figure}[!htb]
	\centering
	\includegraphics[width=0.45\textwidth]{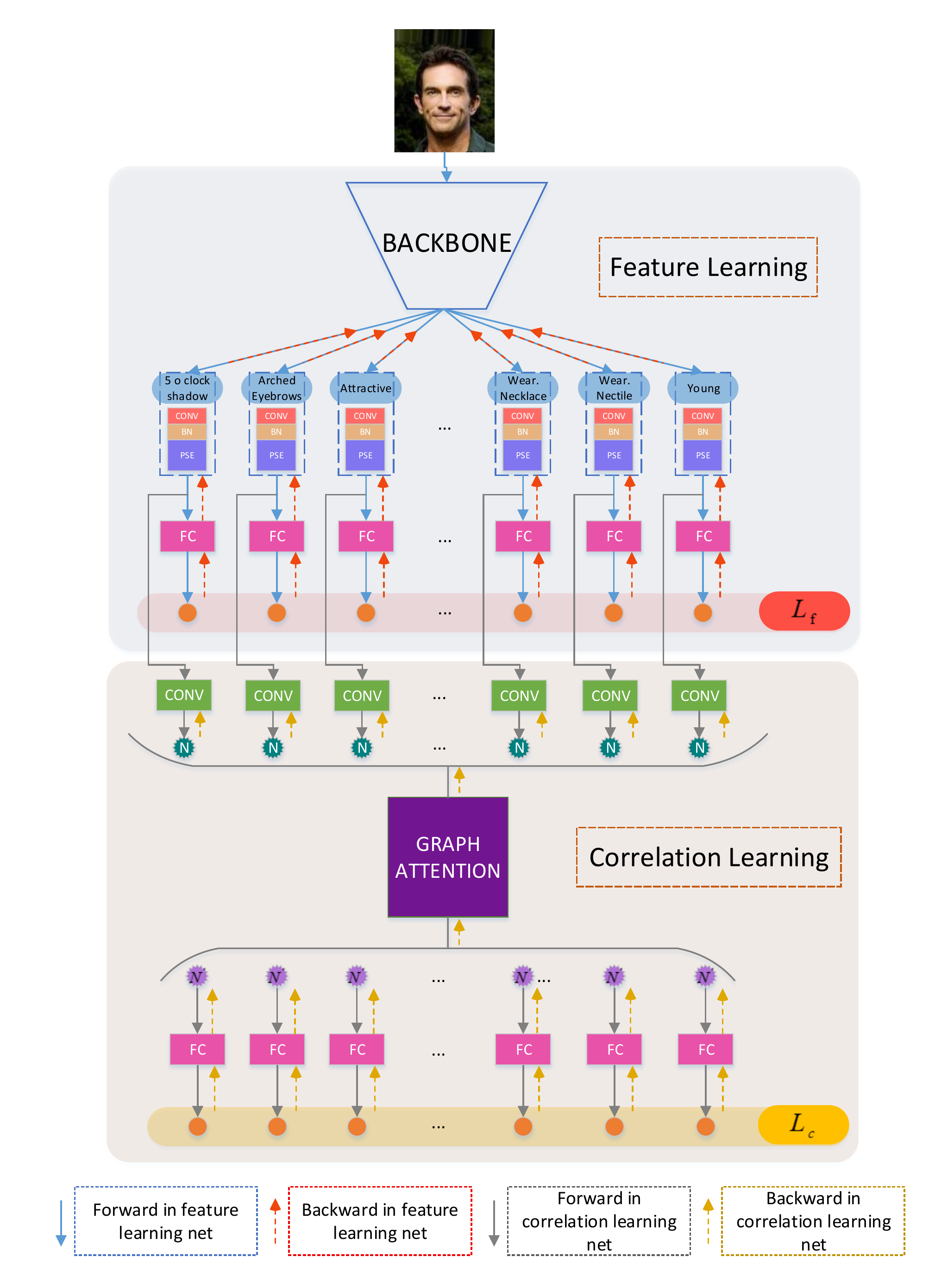}
	\caption{Flowchart of our proposed approach for multiple facial attributes recognition. The whole network consists of the feature learning and the correlation learning network. In the feature learning net, we use $M$ task-specific blocks, the structure of which is exactly the same, to extract the corresponding feature of $M$ attributes. The input of the correlation learning net are the $M$ sets of feature maps from each block in the feature learning net. After the convolution operation, each set of features is treated as a node in graph, and the correlation of the nodes is explored by graph attention machanism. The output of the graph attention are $M$ refined nodes, which are fed to classifiers respectively to recognize the corresponding attribute. The feature learning net and the correlation net are trained with two independent loss $L_f$ and $L_c$, respectively.
	}
	\label{fig:Overview}
\end{figure}

In order to model the correlations among different attributes, we use a graph attention layer (GAL), which is initially proposed to model and infer the relation in knowledge graph in \cite{velickovic2017graph}, to create high level feature representation across different attributes. Our aim is to set up a graph in which each node represents a feature vector for an attribute and the edge linking between two nodes indicates whether they are directly dependent, in other words, they have the correlation with each other. The graph can be set up by the prior knowledge, \emph{e.g.}, the attribute "wavy hair" is negatively correlated with "straight hair" strongly, and "young" is related to "attractiveness" in some extent, hence there should be links between the corresponding nodes. But from prior knowledge, the strength of the link is difficult or even impossible to determine. Our solution is to explore the correlation in a data-driven way, so that both the link and the strength of it are learned from data. Specifically, we use the attention based architecture, in which the similarities among different attribute features are first measured to generate attention weights, and then these weights are used to linearly combine and augment the feature for classification. Note that the input feature of GAL is from the individual branch, so it dose not fully reflect correlation though they have shared low level features. While GAL's outputs consider the relation of attributes and they are more expressive. Since the features before GAL have already shown their ability for classification, we also propose an optimization scheme in which two cross entropy loss functions $L_f$ and $L_c$ are adopted for constraining two different parts of the network, respectively. The idea is to apply the gradient of $L_f$ only for updating the weight in the feature learning network and the gradient of $L_c$ for the weight in the correlation learning network, as is shown in Figure \ref{fig:Overview}. That is to say, $L_c$ is only responsible for finding the high level attributes correlation. As a summarization, we list the contributions of the paper as follows:

\begin{itemize}
\item A GAL structure for multiple facial attributes recognition is proposed. It is a fully data driven approach to explore attributes' correlations.
\item The separate optimization scheme for GAL and lower layers is also designed. Two gradients streams, computed from two loss functions before and after correlation learning layers, are responsible for updating the weight in GAL and lower layers, respectively.
\end{itemize}
The remainder of the paper is constructed as follows: We first discuss the related work in network structure design in MTL, graph neural network, and attribute analysis. This is followed by the introduction of our proposed method. In the last section, we provide the detailed information of our experiments and give the analysis on our work.

\section{Related Works}\label{sec:2}
This section introduces the related works of the paper. There are many works on MTL, and most of the recent ones are built in deep CNN, focusing on the structure design of the network. Since our work intends to use graph neural network to mine the correlations among different attributes, we give brief introduction the works on graph based network, from which we can learn how to construct the neural network so that it can model relations among the attributes. Note that MTL is a general concept of machine learning and it dose not restrict its application in facial attributes recognition.
\subsection{Network Structure Design in MTL}
MTL has been successfully used in many applications of machine learning. The key issue in MTL is to investigate how to design the structure of the network. The simplest way is to share the layers and their parameters, which is first analyzed in \cite{caruna1993multitask}, and it is proved in \cite{baxter1997bayesian} that sharing more parameters can reduced the risk of overfitting. Besides the hard sharing, there can also be soft sharing which means that each task has its own model and parameters, but the distance of parameters is regularized to encourage them to be similar \cite{yang2016trace}. Many deep learning approaches used MTL, explicitly or implicitly, as part of their model, can actually be regarded as either soft or hard parameter sharing. In deep CNN, hard parameters sharing is ordinary, which forces the lower convolution layers to use the same parameters while keeps several task-specific parameters in higher fully connected layers.

Besides soft or hard parameters and layers sharing, there are also flexible ways for the network structure design. \cite{long2015learning} consider the relation of different tasks. In their work, except sharing the lower convolution layer, they place matrix priors in the task-specific fully connected layers, which allow the model to learn the relation between tasks. The matrix prior provides extra constraints about tasks, but it is based on previous knowledge which is not fully data-driven. \cite{lu2017fully} gives a fully adaptive feature sharing method by gradually generating the network branches. They start from a thin network with only one final layer being task-specific, and dynamically widen it by greedily creating more branches with the task-specific parameters. This greedy approach to adapt the task-specific branch is a data-driven way to determine the network structure, but the solution is obviously not optimal, and it is rather time-consuming. \cite{misra2016cross} starts from two identical separate CNN models with different but soft parameters shared between them. They add an extra unit, named the cross stich unit, to share the same level features from each model. The cross stich unit takes the input feature from both models, processes them by simple calculation, and distributes them back into each model. With cross stich units at different levels of the two models, the sharing happens at multiple stage of the feature representations, but the training becomes very unstable, particularly for large number of tasks. \cite{ruder122017sluice} improves \cite{misra2016cross} by taking into the task hierarchy. They allow different levels of the feature to directly give the task-specific layer for final prediction, in other words, the loss gradients can be directly given to lower layers, which makes the training more stable.

Except the network structure design, there are also other issues, such as to determine the weight for the loss of each task, to incorporate auxiliary loss to improve the performance. Due to the page limit, we can not elaborate them.

\subsection{Graph Neural Network}

The idea of Graph Neural Network (GNN) is first proposed in \cite{gori2005new,scarselli2009graph} to deal with graph-structured data. Different from CNN, which is suitable for the data with the regular grid-like structure, \emph{e.g.} image, GNN mainly deals with the data in irregular domain, like social network, 3D meshes, or telecommunication network \emph{et. al.}. \cite{kipf2016semi} introduces the convolution operation onto graph, and proposes the multi-layer Graph Convolution Network (GCN). Similar to the convolution in CNN, the graph convolution also computes a weighted linear combination in its neighbourhood. The key difference is that the neighbourhood of a node is irregular and determined by the edge link between nodes, hence the structure of the graph needs to know before the convolution. They also prove that the convolution on graph can performed easily and equivalently in spectral domain. \cite{velickovic2017graph} present the Graph Attention (GAT) network, in which the graph structure can be totally learned or refined from data by the self-attention mechanism. Moreover, GAT can be computed efficiently without out matrix inversion. In our work, GAT is used to explore the correlations among facial attributes without knowing any prior knowledge on them.

\subsection{Attribute Analysis}
Face has many high level important attributes. The algorithms for single facial attribute, such as gender, age or kinship \cite{levi2015age,ranjan2017hyperface,robinson2016families} usually consider prior knowledge on face, and intend to extract the discriminative feature or loss functions. However, building a single deep model for multiple facial attribute is still difficult. Mainly because too much design on a particular attribute may not generalize to the others. \cite{liu2015deep} propose a CelebA dataset in which amount to 40 binary attributes requires to be estimated by a single model.  \cite{han2017heterogeneous} provides a DMTL (deep MTL) approach in which they first divide attributes into several groups and construct group-specific layer. Then output from these layers are used further by attribute-specific layers. This work actually designs the network structure based on the prior knowledge, and it gives the best performance. \cite{hand2017attributes} uses a similar idea, but they intend to find and use the correlations among different attributes. They design a simple fully connected AUX layer which takes all the attribute-specific feature as input and refine them before making final predictions. Among the above works, only \cite{han2017heterogeneous,hand2017attributes} consider the correlation for attributes. Both of them are highly dependent on prior knowledge. Although the AUX in \cite{hand2017attributes} is a data-driven approach, one single fully connected layer is still not enough for correlation mining.

\section{Our Approach}\label{sec:3}

\noindent In multiple facial attributes recognition, correlation between attributes always exist, and they deserve to be exploited better in a single MTL model. Basically, there are two types of difficuties to consider the correlation of attributes. First, attributes are heterogeneous in many aspects. \emph{E.g.}, "blonde hair" focuses on color while "big lips" describes geometry. Besides, "blonde hair" and "big lips" are both relatively low-level attributes that can be identified directly from the face image, while attribute like "attractiveness" is high level semantic attribute. The relation between each low level attribute and "attractiveness" is difficult to determine. Only when using a data-driven approach, can we reduce subjective influence and get an unbiased cognitive result based on the average aesthetic judgment of the certain dataset. Actually, the heterogeneity of facial attributes determines it is easy to introduce bias with the guidance of prior knowledge, hence the reliable correlation could not be found only with prior knowledge.

In order to fully explore the relationship among $M$ attributes, we divide framework into two parts, as is shown in Figure \ref{fig:Overview}. One part is feature learning network (FLN) with a backbone of several full shared layers, and a total of $M$ task-specific branches. Each branch is of the same structure but different parameters, which is used to extract the unbiased features of $M$ attributes respectively. The other part is the correlation learning network (CLN). Here, we introduce the concept of graph, and regard the $M$ sets of features extracted from the feature learning network as $M$ nodes in graph. The nodes are then given to the GAL to explore the relation among $M$ attributes, and give the weight of each relation to represent the strength of the correlation. Then, by integrating the information from each node, the refined complete feature information can be obtained and the classifier can be learned in an unbiased way.

\subsection{Feature Learning Net}
\begin{figure}[htb]
	\centering

	\includegraphics[width=0.4\textwidth]{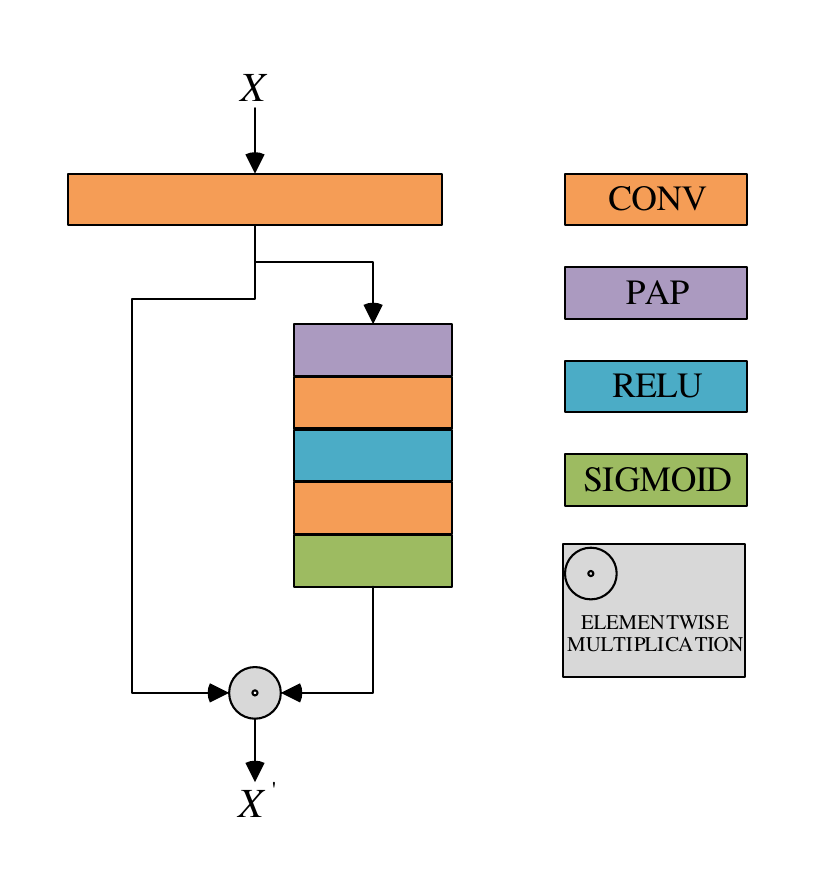}
	\caption{Flowchart of PSE module. PSE module firstly makes a feature compression, calculating the average value of the same position in different channels, which is called position average pooling (PAP), and thus get a single channel feature map with attentioned position. Secondly, PSE module uses two layers of convolution and sigmoid function to activate the position attention feature map, and will obtain a mask which focuses on local information. Finally, PSE computes the elementwise multiplication of the output of the batchnorm layer and the mask. 
	}
	\label{fig:pse}
\end{figure}

We use Alexnet-cvgj model \cite{Simon2016ImageNet}, without the two fully connected layer, as the shared backbone to extract the low level shared features. The output of the shared layer is given to $M$ task-specific learning branches, each branch corresponding to one attribute. The branch consists of a layer of $1\times 1$ convolution, batchnorm and position squeeze excitation (PSE) module, which is proved to be useful for finding the relevant in spatial dimension. The diagram of PSE module can be found in Figure\ref{fig:pse}. The output of one whole branch is one feature set, which has enhanced spatial information. All the $M$ feature sets are fed to their corresponding fully connected layers for classification, respectively. The cross entropy losses from different classifiers are summed for gradients calculation, and then $M$ branches are updated separately, there is no common parameters between each other and thus no interplay among branches. While the shared layer receives the effect from all the $M$ losses, making it eventually learn how to transform the input image to global features, and meet the requirements of $M$ branches at the same time.

\subsection{Correlation Learning Net}
\begin{figure}[ht]
	\centering
	\includegraphics[width=0.45\textwidth]{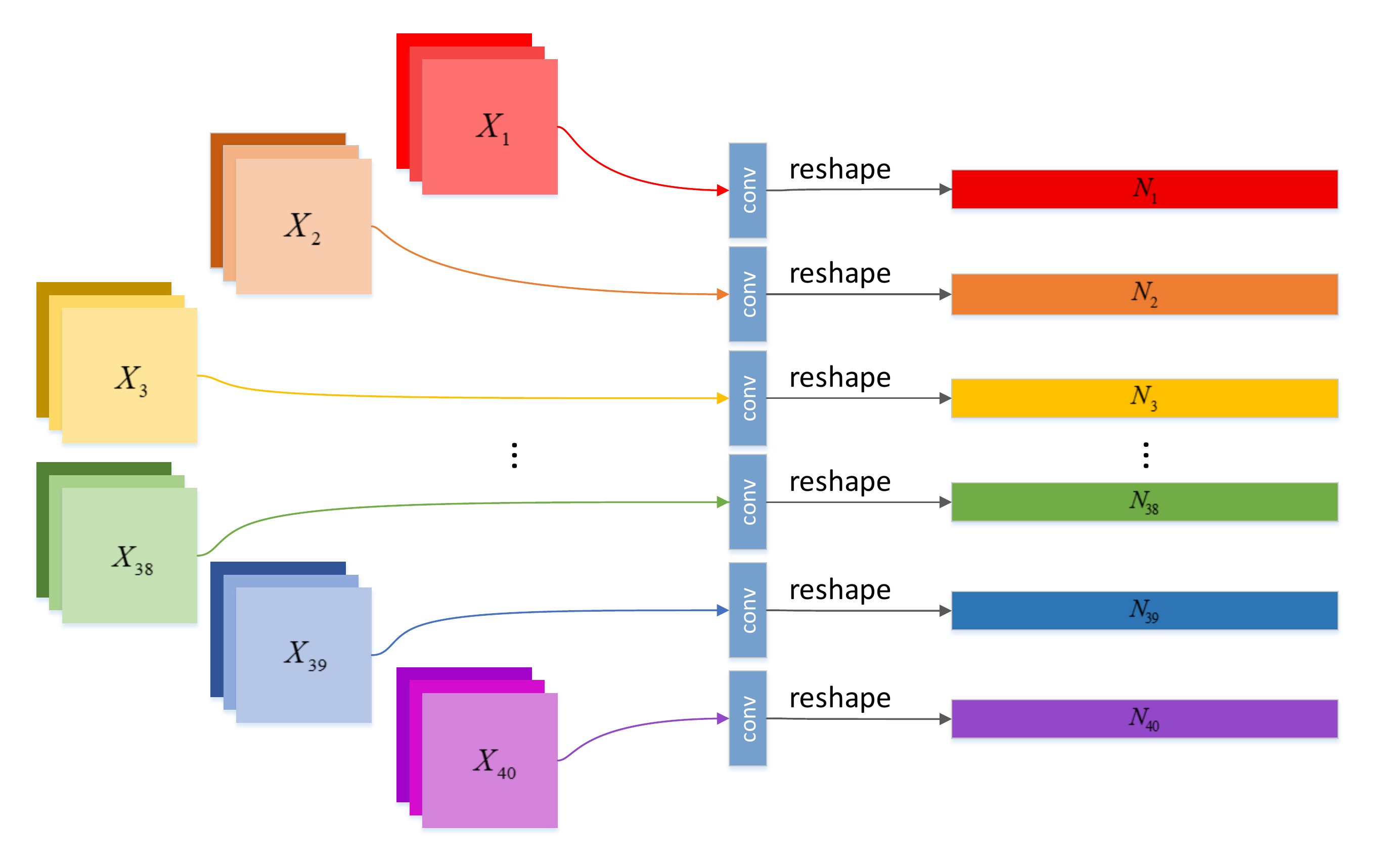}
	\caption{Operation befor GAL. The input feature $\mathbf{X}_{i}$ is the output of the branches in FLN. We consider the $1\times 1$ convolution operation as projection. While the reshape operation makes features into a node, and thus help the learning of the graph attention network.}
	\label{fig:map}
\end{figure}

Correlation Learning Net (CLN) consists of two parts: one is to project them into a certain feature space and add them into the graph node list; the other uses the GAT and find the correlation among the nodes. As is shown in Figure \ref{fig:Overview} and Figure \ref{fig:map}, the output features of the branch in FLN feed into the CLN. In order to obtain sufﬁcient expressive feature for graph attention learning, there should be at least one learnable linear transformation is required, so we use $M$ $1\times 1$ convolutions, parameters of which is not shared, to enhance the expressiveness. In \cite{hand2017attributes}, fully connected layer is used to execute the projection, because fully connected layer can fuse all the features in one feature set. However, for the task of facial attribute recognition that is sensitive to geometric construction, the use of fully connected layers for mapping leads to the loss of spatial information, so we choose convolution operation to execute the projection of feature space. In order to assist the CLN, we flatten the outputs of the $1\times 1$ convolutions to make them into vector format, and we treat these vectors as nodes. In formalization, we define the input of the CLN, \emph{i.e.} the output of the branches in the FLN as $\mathbf{X}_{i} \in \mathbb{R}^{H\times W\times C}$, here $i$ ranges from 1 to $M$, represents the index of attribute. $H$, $W$ and $C$ are the height, width and the number of channels of feature $\mathbf{X}_{i}$, respectively. The operation of convolution is represented as $F_{conv}(.)$, the operation reshape is as $F_{reshape}(.)$ while the ouput node is as $\mathbf{N}_{i} \in \mathbb{R}^{1\times (H\times W\times C)}$. The calculation function is as follows.
\begin{equation}
\mathbf{N}_{i}=F_{reshape}(F_{conv}(\mathbf{X}_{i})),\qquad i=1,2,\cdots,M
\end{equation}

\begin{figure*}[ht]
	\centering
	\includegraphics[width=0.8\textwidth]{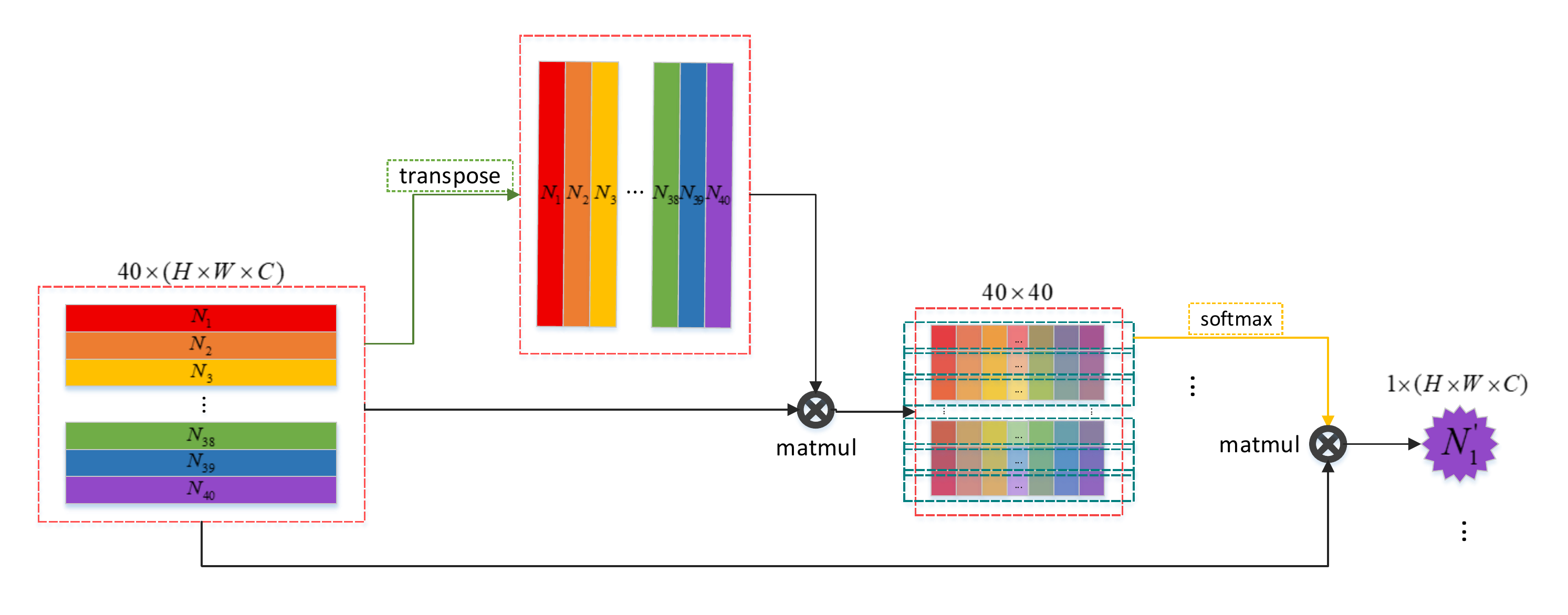}
	\caption{Flow-chart of graph attention (GAT). The input is the processed nodes from mapping operation. We apply the matrix multiplication to compute the affinity matrix among the $M$ nodes. And finally weight the $M$ original nodes according to each row of the affinity matrix to integrade the complementary information from correlated nodes.}
	\label{fig:graph attention}
\end{figure*}

After projection, each node $\mathbf{N}_{i}$ is represented by a multi-dimensional vector with its dimension of $1\times(H\times W\times C)$. As is shown in Figure \ref{fig:graph attention}, We make matrix multiplication between any two vectors. The larger the value is, the higher the similarity and the stronger the correlation will be. The matrix multiplication gives us an $M$ by $M$ affinity matrix $\mathbf{A}$.
\begin{equation}
\mathbf{A}=\mathbf{N}\cdot \mathbf{N}^\mathrm{T},\qquad \mathbf{N}\in \mathbb{R}^{M\times (H\times W\times C)}
\end{equation}
In this matrix $\mathbf{A}$, each row $\mathbf{A}_{i}$ represents the correlation between the corresponding node and all the nodes, including itself.  $\mathbf{A}_{i}$ is then given to softmax to normalize. We then compute the multiplication of the normalized attention and the original $M$ nodes, that is, weight and add all nodes according to the correlation weight, and finally get the refined node $\mathbf{N}^{\prime}$ which integrates all nodes information.
\begin{equation}
\mathbf{N'}_{i}=F_{softmax}(\mathbf{A}_{i})\cdot \mathbf{N},\qquad \mathbf{A}_{i}\in \mathbb{R}^{1\times M}
\end{equation}
As the node vector is the reshape of multiple feature maps, it keeps features' spatial information, therefore, the operation of the matrix multiplication is actually doing the comparison of the similarity between two sets of spatial feature maps. If the two sets of spatial feature maps are highly correlated, the recognition of these two attributes is dependent on a similar source, otherwise, the two attributes do not focus on the similar position. Through the weighted operation, the complementary information of the correlated feature sets is strengthened and the non-correlated of feature sets are suppressed. In conclusion, this data-driven approach allows the data to find the complementary information they need to help with classification on their own.

\subsection{Loss Functions and the Optimization Strategy}
As is shown in Figure\ref{fig:Overview}, there are two loss terms for optimization, $L_f$ and $L_c$ respectively. $L_f$ can be directly computed from each branch of FLN. The prediction of each attribute is evaluated by softmax cross entropy loss first. Then, losses from every branch are summed together to form $L_f$, as is shown in (1).
\begin{equation}
L_f=-\frac{1}{N}\sum_{i=1}^N\sum_{j=1}^M \log p_y^{(i)(j)}
\end{equation}\label{eq:eq1}
\noindent Here $N$ is the total number of the training samples, and $M$ is the attributes number. $y=0\ \mbox{or} \ 1$ is the binary label, and $p_y^{(i)(j)}$ is the estimated value given by the softmax function for the $i$th sample's $j$th attribute. Note that $L_c$ has the same form with $L_f$, but it computes on the refined feature $\mathbf{N}_i'$ after GAL.

When minimizing the two loss terms, the basic idea is to keep each branch task-specific in GAL, meanwhile make each branch after GAL more expressive by considering the correlation of attributes. The basic idea of the optimization strategy is to separate the gradient flow of CLN and FLN. Therefore, the gradient stream from $L_f$ only applies to the parameters in FLN to keep the expressiveness of each branch, and the gradient stream from $L_c$ is only responsible for the parameters update in CLN to make it explore attribute correlations. The gradient flow is shown in Figure\ref{fig:Overview} with dash arrow of different colors.

\section{Experiments}\label{sec:4}
\subsection{Datasets}
We evaluated our method on two challenge face attribute datasets, CelebA \cite{liu2015faceattributes} and LFWA \cite{Wolf2011Effective}.
CelebFaces Attributes Dataset (CelebA) is a large-scale face attributes dataset with about 100 thousand identities and 200 thousand face images. It splits 80 percent images for training, and 20 percent for validation and test. Each image has $M$ attribute annotations, including "5 o'clock shadow", "arched eyebrows", "attractive", "bags under eyes", "bald", "bangs", \emph{etc.}. It provides \emph{In-The-Wild}, \emph{Align and Cropped} sets, and we apply our method on the aligned one.
LFWA dataset contains over 13 thousand images of faces collected from web. It is partitioned into about half for training and half for test. Each image is annotated with exactly the same forty attributes used in CelebA dataset. In order to overcome the overfitting problem, we add some distortion \cite{DBLP:journals/corr/abs-1708-04680} to the training set, and expand the training set to over 75 thousand images.

\subsection{Implementation Details}
The network proposed in this paper has no restriction on the structure of backbone. For simplicity, we use Alexnet-cvgj to build our bottom structure to get global semantic features. To prove the effectiveness of the method we proposed, we provide several contrast test results. We build a network only with FLN as our baseline. Results of joint training of FLN and CLN are marked as GAL-j, and GAL-c. GAL-j means we make joint training on the FLN and the CLN with two terms $L_c$ and $L_f$ loss at the same time. The parameters in FLN and CLN is updated by the gradients of $L_f$ and $L_c$, respectively, which is our proposed optimization scheme. GAL-c refers to that we fix all the parameters in FLN and only train the correlation learning net. In fact, these two cases for training both add constraints to FLN. For our scheme, FLN is required to accurately extract the features of each attribute without deviation, and CLN is required to fully explore the correlation between each attribute pair. If the constraint on FLN is absent, the gradients from the CLN will flow to the FLN to introduce deviation to each independent branch, and then affect the final performance. GAL-j and GAL-c both learn an affinity matrix by data-driven approach. For comparison, we define an affinity matrix artificially, in other words, we build a correlation graph by prior knowledge. This approach is marked as GAL-p. Detailed information of the correlation graph is offered in Table \ref{Custom Pre-classification of Nodes}. There are 8 groups in total, and each group is decided according to the naturally appearing location of attributes. All the nodes in the same group are adjacent, and the sum of values on the edges is one, while different groups have no linking edge.

To assist training, we use the publically available Alexnet-cvgj pretrain model on ImageNet to initialize the shared layers. For all the training images, we first standardize them to $256\times 256\times 3$ size, and then randomly left-or-right flip the images in an online way, before they are fed into the network. On the CelebA dataset, for all the net, we set the initial learning rate to be 0.005, and it will follow a polynomial decay function with the training process going on. Batch size is 256, the max iteration step is 25600. As for baseline, the weight decay is set to be 0.0005, while for other three methods, it becomes 0.001. When train on the LFWA dataset, we apply the cyclical learning rate \cite{7926641} to train all the net. The maximum learning rate is 0.005, minmun is 0, stepsize equals 5000, and has no decay. The total training step is 10000 iteration. 

\subsection{Results and Analysis}
Based on the methods described above, our results on CelebA dataset is listed in Table \ref{res on celebA}, and LFWA is listed in Table \ref{comp on lfwa}. Comparison with results from other current methods is listed in Table \ref{comp on celebA} as well.
\begin{table}[h]
	\centering
	\caption{Custom Pre-classification of Nodes}
	\label{Custom Pre-classification of Nodes}
	\begin{tabular}{|c|p{6cm}|}
		\hline
		\textbf{Group} & \textbf{Attributes}                                                                                                           \\ \hline
		\multirow{3}*{Global}      & Attractive, Blurry, Chubby, Heavy Makeup, Male, Oval Face, Pale Skin, Smiling, Young                                 \\ \hline
		\multirow{3}*{Hair}      & Bald, Bangs, Black Hair, Blond Hair, Brown Hair, Gray Hair, Receding Hairline, Straight Hair, Wavy Hair, Wearing Hat \\ \hline
		\multirow{3}*{Eye}      & Arched Eyebrows, Bags Under Eyes, Bushy Eyebrows, Eyeglasses, Narrow Eyes                                            \\ \hline
		Nose      & Big Nose, Pointy Nose                                                                                                \\ \hline
		\multirow{2}*{Cheek$\&$Ear}      & High Cheekbones, Rosy Cheeks, Sideburns, Wearing Earrings                                                            \\ \hline
		\multirow{3}*{Mouse}      & 5 o'Clock Shadow, Big Lips, Mouth Slightly Open, Mustache, Wearing Lipstick                                          \\ \hline
		Chin      & Double Chin, Goatee, No Beard                                                                                        \\ \hline
		Neck      & Wearing Necklace, Wearing Necktie                                                                                    \\ \hline
	\end{tabular}
\end{table}

\begin{table}[h]
	\centering
	\caption{Results on CelebA Dataset}
	\label{res on celebA}
	\begin{tabular}{lcccc}
		\hline
		Approach         & Baseline & GAL-c & GAL-j          & GAL-p \\ \hline
		Accuracy & 90.73    & 90.89 & \textbf{91.43} & 90.13 \\ \hline
	\end{tabular}
\end{table}

\begin{table}[h]
	\centering
	\caption{Comparison on LFWA Dataset}
	\label{comp on lfwa}
	\begin{tabular}{cp{1cm}p{1cm}cc}
		 \hline
		\multirow{2}*{Approach} & LNets-ANet & MCNN-AUX & 	\multirow{2}*{Baseline} & 	\multirow{2}*{GAL-j} \\ \hline
		Accuracy & 84   & \textbf{86.3}             & 85.19    & 85.25      \\ \hline
	\end{tabular}
\end{table}

\begin{table}[!htb]
	\centering
	\caption{Comparison on CelebA Dataset}
	\label{comp on celebA}
	\begin{tabular}{|l|c|c|c|c|}
		\hline
		& \rotatebox{90}{MCNN-AUX}       & \rotatebox{90}{DMTL}           & \rotatebox{90}{AFFACT}         & \rotatebox{90}{GAL-j}          \\ \hline
		5 Shadow        & 94.51          & \textbf{95.00} & 94.21          & 94.80          \\
		Arched Eyebrows & 83.42          & \textbf{86.00} & 82.12          & 84.16          \\
		Attractive      & 83.06          & \textbf{85.00} & 82.83          & 82.89          \\
		Bags Un Eyes    & 84.92          & 85.00          & 83.75          & \textbf{85.29} \\
		Bald            & 98.90          & 99.00          & \textbf{99.06} & 98.90          \\
		Bangs           & 96.05          & \textbf{99.00} & 96.05          & 96.13          \\
		Big Lips        & 71.47          & \textbf{96.00} & 70.88          & 71.81          \\
		Big Nose        & 84.53          & \textbf{85.00} & 83.82          & 84.35          \\
		Black Hair      & 89.78          & \textbf{91.00} & 90.32          & 90.34          \\
		Blonde Hair     & 96.01          & \textbf{96.00} & 96.07          & 95.98          \\
		Blurry          & 96.17          & 96.00          & 95.50          & \textbf{96.22} \\
		Brown Hair      & 89.15          & 88.00          & \textbf{89.16} & 89.15          \\
		Bushy Eyebrows  & 92.84          & 92.00          & 92.41          & \textbf{92.88} \\
		Chubby          & 95.67          & \textbf{96.00} & 94.98          & 95.75          \\
		Double Chin     & 96.32          & \textbf{97.00} & 96.18          & 96.40          \\
		Eyeglasses      & \textbf{99.63} & 99.00          & 99.61          & 99.55          \\
		Goatee          & 97.24          & \textbf{99.00} & 97.31          & 97.40          \\
		Gray Hair       & 98.20          & 98.00          & 98.28          & \textbf{98.34} \\
		Heavy Makeup    & 91.55          & \textbf{92.00} & 91.10          & 91.80          \\
		H. Cheekbones   & 87.58          & \textbf{88.00} & 86.88          & 87.98          \\
		Male            & 98.17          & 98.00          & \textbf{98.26} & 98.17          \\
		Mouth S. O.     & 93.74          & \textbf{94.00} & 92.60          & 93.92          \\
		Mustache        & 96.88          & \textbf{97.00} & 96.89          & 96.83          \\
		Narrow Eyes     & 87.23          & \textbf{90.00} & 87.23          & 87.57          \\
		No Beard        & 96.05          & \textbf{97.00} & 95.99          & 96.21          \\
		Oval Face       & 75.84          & \textbf{78.00} & 75.79          & 75.78          \\
		Pale Skin       & 97.05          & 97.00          & 97.04          & \textbf{97.22} \\
		Pointy Nose     & 77.47          & \textbf{78.00} & 74.83          & 77.61          \\
		Reced. Hairline & 93.81          & \textbf{94.00} & 93.29          & 93.76          \\
		Rosy Cheeks     & 95.16          & \textbf{96.00} & 94.45          & 95.22          \\
		Sideburns       & 97.85          & \textbf{98.00} & 97.83          & 97.93          \\
		Smiliing        & 92.73          & \textbf{94.00} & 91.77          & 92.98          \\
		Straight Hair   & 83.58          & \textbf{85.00} & 84.10          & 83.67          \\
		Wavy Hair       & 83.91          & \textbf{87.00} & 85.65          & 84.32          \\
		Wear. Earrings  & 90.43          & \textbf{91.00} & 90.20          & 90.34          \\
		Wear. Hat       & \textbf{99.05} & 99.00          & 99.02          & \textbf{99.05} \\
		Wear. Lipstick  & 94.11          & 93.00          & 91.69          & \textbf{94.18} \\
		Wear. Necklace  & 86.63          & \textbf{89.00} & 87.85          & 86.96          \\
		Wear. Necktie   & 96.51          & \textbf{97.00} & 96.90          & 96.62          \\
		Young           & 88.48          & \textbf{90.00} & 88.66          & 88.57          \\ \hline
		Average         & 91.29          & \textbf{92.60} & 91.01          & 91.43          \\ \hline
	\end{tabular}
\end{table}

\subsection{Analysis on GAL}
As is listed in Table \ref{res on celebA}, the mean accuracy of Baseline, GAL-c, GAL-j, and GAL-p are 90.73$\%$, 90.89$\%$, 91.43$\%$ and 90.13$\%$ respecitively. Obviously, adding CLN do help increase the performance., and the proposed optimization strategy plays an important role in facial attributes recognition task as well. GAL-c train the FLN and the CLN separately. Parameters in the FLN are loaded from the Baseline model, which is regarded as the best attribute feature extractor. During the training process, parameters in the FLN are fixed, only the parameters in the correlation learning net can be updated. GAL-j train the FLN and the CLN with two separate gradient streams from $L_c$ and $L_f$ at the same time. Actually, GAL-c and GAL-j both constrain the attribute feature independency and the attribute correlation, intending to fully extract independent feature and fully exploit correlation among attributes. But the mean accuracy of GAL-j is 0.54$\%$ higher than that of GAL-c. The difference between the two methods lies in whether the FLN and the CLN share the training process. We believe that with the guidance from FLN and our proposed optimization strategy, CLN will not easily fall into local minima and can jump out of local minima as training goes on. Moreover, with a fixed FLN , as is shown by the GAL-c training results, the final model cannot reach the optimum. The training method of GAL-p is exactly the same with GAL-j, but GAL-p's affinity matrix is defined artificially, GAL-j learns the affinity matrix on its own. It is quite reasonable that the mean accuracy of GAL-p is even lower than baseline as it is very hard to determine the adjacency of attributes and the degree of correlation. For the dataset of LFWA, the mean accuracy of GAL-j is 0.06$\%$ higher than the Baseline, as is listed in Table \ref{comp on lfwa}, indicates that GAL is effective, but the effect is not obvious. The main reason is LFWA dataset is too small, so it is difficult to train our model just using LFWA.

\subsection{Comparison with Other Approaches}
As is shown in Table \ref{comp on celebA} and Table \ref{comp on lfwa}, performance of our approach is better than AFFACT \cite{DBLP:journals/corr/GuntherRB16} and LNets-ANet \cite{liu2015deep}. DMTL \cite{han2017heterogeneous} method has the best performance at present. DMTL also explores the correlation of attributes. It learns the common shared features first, and uses several branches to learn different groups of features. Finally, individual attribute classification is made based on its group features. The classification of the group is determined by the prior knowledge. The training of DMTL takes at least 100,000 iterations, and need to pretrain its model on CASIA dataset. While our net is trained in end-to-end method, we can get results in 25600 iterations and have no need to pretrain on a much bigger dataset. MCNN-AUX \cite{hand2017attributes} is an end-to-end network, it uses fully connected layers to explore the correlation of attributes. As it is a quite shallow network, on LFWA it suffers less overfitting than our approach. But our approach has better performance on the large dataset CelebA.

\subsection{Visualizations}
\begin{figure}[ht]
	\centering
	\includegraphics[width=0.45\textwidth]{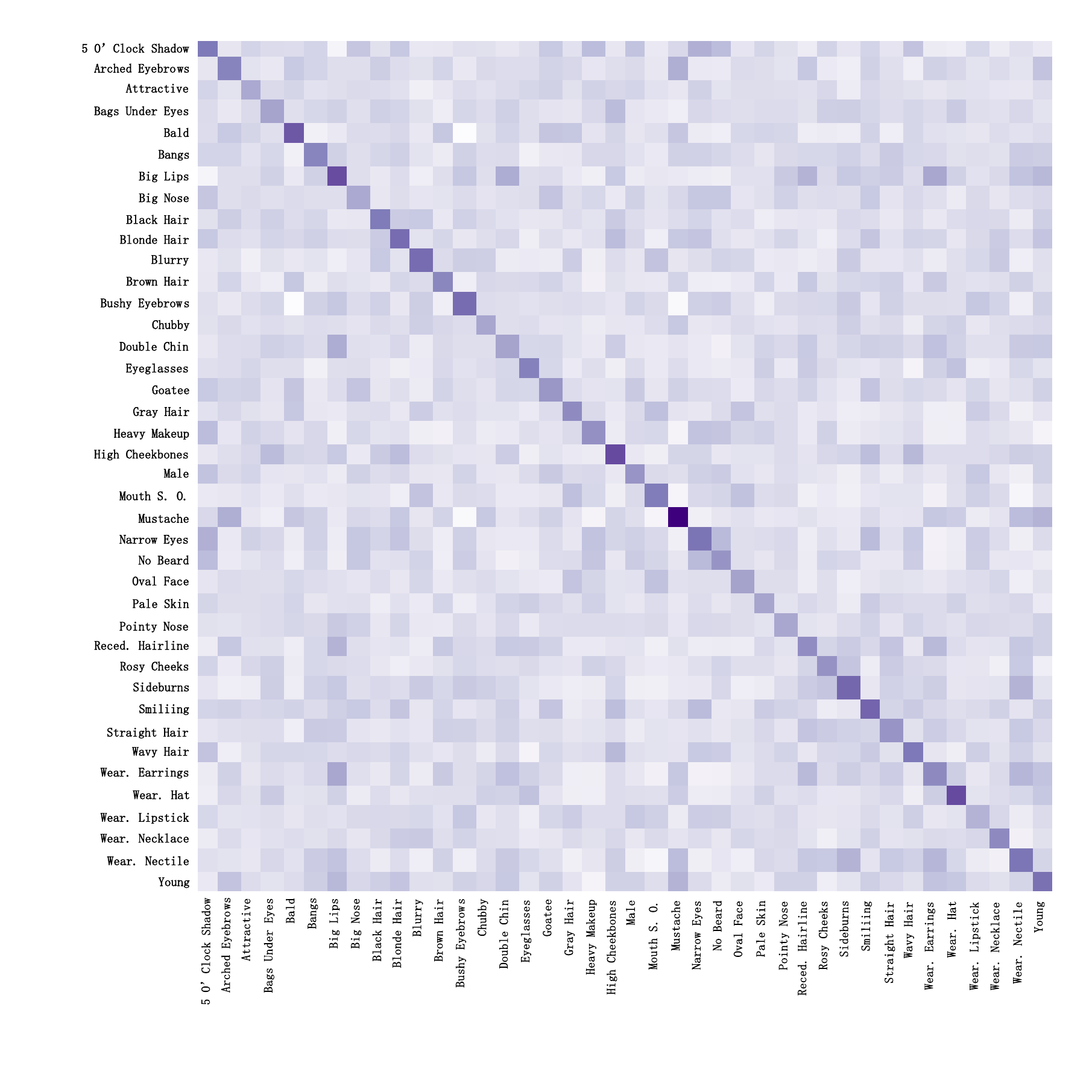}
	\caption{Heatmap of the affinity matrix in GAL-j. We have all the $M$ attributes on the x-axis and the y-axis. The deeper the color is, the stronger relationship exists in the two attributes. Best viewed in color.}
	\label{fig:affinity matrix}
\end{figure}
Figure \ref{fig:affinity matrix} shows the heatmap of affinity matrix learned in GAL-j. We can clearly see that every attribute has the strongest correlation with itself. Besides, there are some intuitive relationships in the heatmap, such as "smile" and "narrow eyes". With the data-driven approach, we can also find some interesting point like "young" and "wearing earings", "lipstick", "hat" and "necktie" are correlated strongly, which indicating the subjective concept of young. We noticed that "high cheekbones" and "smile" is closely related. Actually, when someone smile, he is more likely to be seen as having high cheekbones, may be noise label is introduced on this reason. In addition, there are some attribute almost have nothing to do with each other. For instance, "bush eyebrows" has no relation with "bags under eyes" and "mustache", and "mustache" is absolutely not related to heavy makeup.

\section{Conclusion}\label{sec:5}
In order to fully explore the relations among attributes and synthesize the information of related attribute features, we propose to add GAL to the FLN to study the independent features and correlation at the same time in an end-to-end way. After the FLN, we use the convolution layer to reduct dimension and reshape the independent feature, making it a node. The $M$ nodes are fed into the GAL layer and the relations among them is mapped with a data-driven approach. From the visualization of the affinity matrix, we can see that the intuitively related attributes are still correlated in the graph, but the correlation degree, that is, the correlation weight is determined. Meanwhile, the affinity matrix learned from the data also gives us some unexpected attribute relations. In the process of training, we found that training with two independent loss simultaneously can guide the network to find the optimal, prevent it from falling into the local minimum point. Experiments show this method can obtain a good classification result. All in all, the approach we propose is effective and meets our expectations. You can find our codes on \url{https://github.com/crazydemo/facial-attribute-classification-with-graph}

\bigskip
\bibliographystyle{aaai}
\bibliography{main}
\end{document}